\pdfoutput=1

\documentclass[11pt]{article}

\usepackage[final]{acl}

\usepackage{times}
\usepackage{latexsym}
\usepackage{graphicx}
\usepackage{multirow}
\usepackage{hyperref}
\usepackage{xurl}

\usepackage[T1]{fontenc}

\usepackage[utf8]{inputenc}

\usepackage{microtype}

\title{IndiText Boost: Text Augmentation for Low Resource India Languages}

\author{Onkar Litake\thanks{* indicates equal contribution} , Niraj Yagnik\footnotemark[1] \and Shreyas Labhsetwar\footnotemark[1] \\
        Department of Computer Science and Engineering \\
        University of California, San Diego \\
        9500 Gilman Drive, La Jolla, CA 92093, USA \\
        \texttt{\{olitake, nyagnik, slabhsetwar\}@ucsd.edu}
}

\begin{document}
\maketitle
\begin{abstract}
Text Augmentation is an important task for low-resource languages. It helps deal with the problem of data scarcity. A data augmentation strategy is used to deal with the problem of data scarcity. Through the years, much work has been done on data augmentation for the English language. In contrast, very less work has been done on Indian languages. This is contrary to the fact that data augmentation is used to deal with data scarcity. In this work, we focus on implementing techniques like Easy Data Augmentation, Back Translation, Paraphrasing, Text Generation using LLMs, and Text Expansion using LLMs for text classification on different languages. We focus on 6 Indian languages namely: Sindhi, Marathi, Hindi, Gujarati, Telugu, and Sanskrit. According to our knowledge, no such work exists for text augmentation on Indian languages. We carry out binary as well as multi-class text classification to make our results more comparable. We get surprising results as basic data augmentation techniques surpass LLMs.
\end{abstract}

\section{Introduction}

Data Augmentation is a process in NLP that enables us to artificially increase the training data size by generating various versions of the real data. Different downstream tasks require different data augmentation methods. We focus on data augmentation for text classification. To perform better on the classification task, the data must be changed such that it maintains the class categories. 

The size of the training data is expanded through data augmentation, which enhances model performance. The model performs better the more data we have. The distribution of the generated augmented data should neither be too similar nor too different from the original. 

The problem of data scarcity is addressed using the data augmentation strategy. A lot of work has been done over the years on data augmentation for the English language. Indian languages, on the other hand, have received very little research. 

In this project, we are going to work on data augmentation for Indian languages. This project stems from the motivation that no such work is carried out in the past. We are going to implement existing approaches for data augmentation like Easy Data Augmentation, Back Translation, Paraphrasing, Document Generation, etc. We plan on making some changes to these existing methods to tailor them specifically for Indian languages. 

This work will be helpful in a lot of NLP tasks like News Classification, Hate Detection, Emotion Analysis, Sentiment Analysis, and Spam Classification. All these tasks currently face the problem of data scarcity, which will be solved by our project. We aim to provide different methods for Data Augmentation. The user can implement different techniques and select what works best for the given use case. Hence the outcome of our research work is:
\begin{itemize}
    \item Implementing different data augmentation techniques.
    \item Comparing all the methods to establish a baseline.

\end{itemize}

\section{Related Work}

Data augmentation has been extensively researched in the field of Natural Language Processing (NLP) as a valuable approach to address the lack of training data \cite{feng2021survey}. One prominent technique, introduced by \cite{sennrich2015improving}, is back translation, which enhances the performance of machine translation (MT) measured by BLEU scores \cite{papineni2002bleu}. This method involves translating sentences into a different language and then translating them back into the original language, thereby augmenting the original text.

\cite{fadaee2017data} present a data augmentation method specifically designed for low-frequency words. This method generates new sentence pairs that include these infrequently occurring words. \cite{kafle2017data} contribute two data augmentation techniques for enhancing visual question answering. The first approach employs semantic annotations to augment the questions, while the second technique utilizes an LSTM network \cite{hochreiter1997long} to generate new questions from images. \cite{wang2015s} propose an augmentation technique that involves replacing query words with their synonyms. The synonyms are retrieved based on cosine similarities calculated using word embeddings. Additionally, \cite{kolomiyets2011model} propose a data augmentation strategy that involves replacing temporal expression words with their corresponding synonyms. This approach relies on the vocabulary provided by the Latent Words Language Model (LWLM) and WordNet.

\cite{csahin2019data} suggest two text augmentation methods that rely on dependency trees. The initial technique involves cropping sentences by removing specific dependency links. The second technique involves rotating sentences using tree fragments that pivot around the root. \cite{chen2020mixtext} propose a text augmentation approach that involves interpolating input texts within a hidden space. \cite{wang2018switchout} propose a method for augmenting sentences by randomly substituting words in both input and target sentences with vocabulary words. SeqMix \cite{guo2020sequence} proposes a technique for generating augmentations by smoothly merging input and target sequences.

EDA \cite{wei2019eda} employs four operations to create data augmentation: synonym replacement, random insertion, random swap, and random deletion. In a different approach, \cite{kobayashi2018contextual} suggests stochastically replacing words with predictions generated by a bi-directional language model. \cite{andreas2019good} proposes a compositional data augmentation technique that constructs synthetic training examples by substituting text fragments in a real example with other fragments appearing in similar contexts. \cite{kumar2020data} utilize pretrained Transformer models, such as GPT-2, BERT, and BART, for conditional data augmentation. They feed the concatenation of class labels and input texts into these models to generate augmented texts. Additionally, \cite{kumar2020data} propose a language model-based data augmentation method. This method involves fine-tuning a language model on a limited training dataset and then using class labels as input to generate augmented sentences. \cite{min2020syntactic} explore various syntactically informative augmentation techniques by applying syntactic transformations to original sentences and demonstrate that subject/object inversion can enhance robustness to inference heuristics.

\section{Dataset}
The proposed work obtains data augmentation for a total of the following 6 languages:
\begin{itemize}
    \item Sindhi \cite{sind}: This dataset is a valuable resource for researchers in the field of Sindhi Natural Language Processing (NLP), as it represents one of the limited publicly accessible collections of Sindhi articles. It comprises a total of 3364 articles that span three distinct categories: sports, entertainment, and technology. The dataset was sourced from awamiawaz.pk, and its availability greatly facilitate investigations and advancements in Sindhi NLP research.

    \item Hostility Detection Dataset in Hindi \cite{bhardwaj2020hostility}: The dataset contains ~8200 hostile and non-hostile texts from social media platforms like Twitter, Facebook, and WhatsApp. The hostile class is further subdivided into Fake, Offensive, Hate, and Defamation.

    \item L3Cube-MahaHate - Marathi \cite{velankar2022l3cubemahahate}: Dataset consists of over 25000 distinct tweets labeled into four major classes i.e hate, offensive, profane, and not-hate.

    \item Gujarati News Dataset - Gujarati \cite{arora-2020-inltk}: his data set contains ~6500 news article headlines which are collected from Gujarati news websites. It contains 3 labels namely entertainment, business, and technology.

    \item Multi domain corpus for sentimental analysis - Telugu 2 Class \cite{gangula-mamidi-2018-resource}: It contains 339 different Telugu song lyrics written in Telugu script. Out of them, 230 are positive and 109 are negative. It contains a total of 13997 sentences and 81798 words.

    \item Telugu NLP - Telugu 5 Class \cite{telugu_nlp}: The data is extracted from Telugu book. It contains the following labels: business, editorial, entertainment, nation, and sport.

\end{itemize}

For each of the six languages, the proposed work performs data augmentations for two tasks: i) Binary Classification and ii) Multi-Class Text Classification. Thus a total of 12 datasets are collected for the research and experimentations. One hundred examples are sampled from each dataset as the train set. The reason behind selecting the lower number of sentences is to replicate the scenario of implementing the augmentation techniques on low-resource Indian languages, which won't have sufficient labeled data. 

For languages that did not have a dataset for binary text classification, our work filters the multi-class dataset to include only two labels while maintaining class balance.

\section{Baselines}

For the baseline, for each language, we fine-tune the pre-trained BERT model (more specifically, the bert-base-multilingual-cased model) for the un-augmented data. The proposed work adopts a standard set of hyperparameters for each experiment. The list of hyperparameters for each model with the number of epochs being 3 and the max\_length of 512.

The accuracy scores obtained by the fine-tuned BERT model for the augmented data of each language are recorded as the benchmark for all the augmentation techniques the work intends to utilize. 

\section{Methodology}

The proposed work experiments with a series of data augmentation techniques to understand what is the most efficient method. The general pipeline to test the augmentation technique remains standard across all the methods and is illustrated in \ref{fig:method-test}.Each augmentation task is provided with an initial 100 sets of text samples for each classification task and each language. We generate 1 augmentation for each sentence, hence a total of 100 augmented examples are generated. Post-generation of the new text samples, concatenation is performed to get the augmented text database. This augmented text is now used to fine-tune the BERT model as used by the baseline models. The pre and post-augmentation accuracy scores are compared to deduce the efficacy of the augmentation technique. A deep dive into each augmentation technique is provided:

\begin{figure*}[th]
  \centering
  \includegraphics[width=\linewidth]{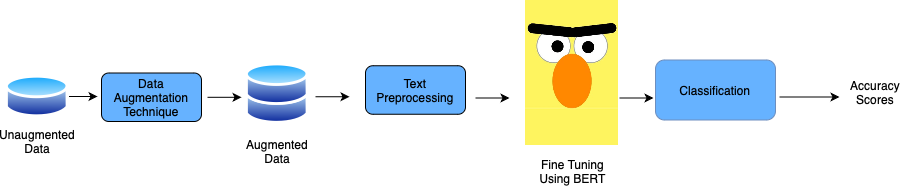}
  \caption{Finetuning BERT for augmented data}
  \label{fig:method-test}
\end{figure*}

\vspace{0pt}

\subsection{Easy Data Augmentation(EDA)}

\cite{wei2019eda} : The work first proposes to leverage EDA to provide a simple yet effective solution for generating augmented data. The technique aims to improve the generalization and efficacy of Machine Learning models by allowing the model to be trained on a diverse variety of the original day. The EDA method consists of four basic operations that can be applied to individual data samples and the proposed work intends to apply these techniques to Indian Languages:

\begin{itemize}

\item \textbf{Synonym Replacement:} For this technique, we will randomly select words from the input text of an Indian language and replace them with their corresponding synonyms. This technique will thus introduce semantic variations in the new augmented text while maintaining the meaning of the data. N words are randomly selected from the input sentence (that are not stop words); these words and then replaced by their synonyms. 

        \item \textbf{Random Insertion:} This technique will find a random synonym of a random word in the sentence written in the Indian language (that is not a stop word). This synonym is inserted at an arbitrary position in the sentence.

        \item \textbf{Random Swap (RS)}: Randomly choose two words in the sentence and swap their positions. Do this n times.
        
        \item \textbf{Random Deletion (RD): }This technique randomly removes words in the text with a certain probability (p). The model is forced to rely on the remaining words after certain terms are eliminated, possibly leading to the learning of more robust representations.

\end{itemize}

\subsection{Back-Translation:}

\begin{figure}[th]
  \centering
  \includegraphics[width=\linewidth]{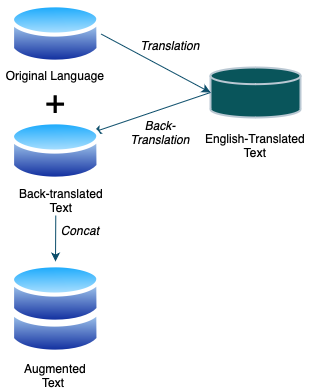}
  \caption{Backtranslation Technique}
  \label{fig:back-trans}
\end{figure}

\cite{sennrich2015improving} In this technique, the input text is translated from one language to another and is retranslated back to the same language. Upon back-translation, the back-translate sentences are mixed with the source sentences to train the model on augmented data.The illustation for back-translation is provided in \ref{fig:back-trans}.

The work utilizes the google translate API \cite{googletranslateapi} for the translation and back-translation tasks.

\subsection{Paraphrasing: }
\begin{figure}
  \centering
  \includegraphics[width=\linewidth]{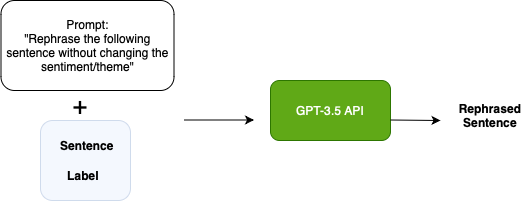}
  \caption{Paraphrasing Module}
  \label{fig:para}
\end{figure}

\cite{andreas2019good} With this technique, the proposed work employs Generative Large Language Models to rephrase the original input text while preserving the underlying context and meaning. More specifically, the GPT3.5 API \cite{openai_gpt35} with the DaVinci by OpenAI is utilized for generating the paraphrased sentences. Each of the hundred text samples, along with their label, is used as the input to the prompt as indicated in \ref{fig:para}. The langdetect library \cite{langdetect} ensures the rephrased text complies with the language and length requirements.

\subsection{Text Extender: \cite{kumar2020data}} In this technique, the input size is extended by using Language Models (GPT2 specifically) while maintaining the overall context and sentiment of the input. By adding more context and possibly offering fresh viewpoints or facts, this strategy helps lengthen the input.  

\subsection{Text Generation: } \cite{dai2023auggpt} The text Generation module, similar to the paraphraser, utilizes the OpenAI GPT3.5 API \cite{openai_gpt35} and relies heavily on the prompt provided to the API. For the task of generating text similar to the input text and the associated label, a static prompt is provided to the API. The prompt offers an example text sentence associated with the label and asks the model to generate a text similar to the input text sentence. This process is repeated for each text sample on the original text dataset for each language.

All the proposed methods were successfully implemented and tested. The work utilizes Google Colab to conduct the experiments and make optimal use of the GPU. 
All the data augmentation techniques are applied for each classification task and language. The BERT model is fine-tuned for every augmented dataset obtained to make the comparisons.

\begin{table*}
\centering
\begin{center}
\begin{tabular}{ lllllll } 
\hline
\textbf{Language} & \textbf{Methodology} & \textbf{Accuracy} & \textbf{Precision}	& \textbf{Recall} & \textbf{F1 Macro} & \textbf{F1 Mirco} \\

\hline
\multirow{3}{4em}{Sanskrit} & Baseline & 0.53 & 0.265 & 0.5 & 0.346 & 0.53 \\ 
& Synonym Replacement & 0.73 & 0.73 & 0.73 & 0.73 & 0.729 \\ 
& Random Insertion & 0.62 & 0.618 & 0.616 & 0.617 & 0.62 \\ 
& Random Swap & 0.67 & 0.682 & 0.659& 0.654 & 0.67 \\ 
& Random Delete & 0.73 & 0.731 & 0.7322 & \textbf{0.729} & 0.73 \\ 
& LLM Expand & 0.67 & 0.676 & 0.674 & 0.669 & 0.67 \\ 
& Back-Translate & 0.71 & 0.718 & 0.7023 & 0.7104 & 0.71 \\ 
& Paraphrase & 0.7 & 0.698 & 0.697 & 0.698 & 0.7 \\ 
& Generate & 0.617 & 0.624 & 0.617 & 0.621 & 0.617 \\

\hline
\multirow{3}{4em}{Marathi} & Baseline & 0.656 & 0.664 & 0.656 & 0.652 & 0.656 \\ 
& Synonym Replacement & 0.666 & 0.693 & 0.666 & 0.678 & 0.667 \\ 
& Random Insertion & 0.686 & 0.701 & 0.686 & 0.693 & 0.687 \\ 
& Random Swap & 0.6862 & 0.686 & 0.686 & 0.686 & 0.686 \\ 
& Random Delete & 0.6568 & 0.657 & 0.656 & 0.656 & 0.656 \\ 
& LLM Expand & 0.696 & 0.696 & 0.696 & 0.696 & 0.696 \\ 
& Back-Translate & 0.725 & 0.738 & 0.725 & \textbf{0.732} & 0.725 \\ 
& Paraphrase & 0.705 & 0.717 & 0.7058 & 0.711 & 0.705 \\ 
& Generate & 0.715 & 0.719 & 0.715 & 0.717 & 0.715 \\

\hline
\multirow{3}{4em}{Hindi} & Baseline & 0.91 & 0.910 & 0.91 & 0.909 & 0.91 \\ 
& Synonym Replacement & 0.94 & 0.940 & 0.94 & \textbf{0.940} & 0.94 \\ 
& Random Insertion & 0.9 & 0.9 & 0.9 & 0.9 & 0.9 \\ 
& Random Swap & 0.91 & 0.9101 & 0.91 & 0.909 & 0.91 \\ 
& Random Delete & 0.88 & 0.880 & 0.88 & 0.879 & 0.88 \\ 
& LLM Expand & 0.91 & 0.914 & 0.91 & 0.909 & 0.91 \\ 
& Back-Translate & 0.9 & 0.9 & 0.9 & 0.9 & 0.9 \\ 
& Paraphrase & 0.91 & 0.911 & 0.909 & 0.910 & 0.91 \\ 
& Text Generate & 0.92 & 0.9206 & 0.919 & 0.920 & 0.92 \\

\hline
\multirow{3}{4em}{Gujarati} & Baseline & 0.528 & 0.542 & 0.53 & 0.535 & 0.528 \\ 
& Synonym Replacement & 0.85 & 0.853 & 0.85 & \textbf{0.851} & 0.85 \\ 
& Random Insertion & 0.84 & 0.854 & 0.84 & 0.847 & 0.839 \\ 
& Random Swap & 0.5 & 0.25 & 0.5 & 0.333 & 0.5 \\ 
& Random Delete & 0.85 & 0.853 & 0.85 & 0.849 & 0.85 \\ 
& LLM Expand & 0.83 & 0.831 & 0.83 & 0.829 & 0.83 \\ 
& Back-Translate & 0.71 & 0.766 & 0.71 & 0.737 & 0.71 \\ 
& Paraphrase & 0.66 & 0.69 & 0.66 & 0.678 & 0.66 \\ 
& Text Generate & 0.7 & 0.705 & 0.7 & 0.7 & 0.7 \\

\hline
\multirow{3}{4em}{Telugu} & Baseline & 0.54 & 0.212 & 0.5 & 0.352 & 0.51 \\ 
& Synonym Replacement & 0.6 & 0.601 & 0.6 & 0.6 & 0.6 \\ 
& Random Insertion & 0.5 & 0.25 & 0.5 & 0.333 & 0.5 \\ 
& Random Swap & 0.68 & 0.673 & 0.666 & 0.637 & 0.67 \\ 
& Random Delete & 0.72 & 0.793 & 0.728 & \textbf{0.772} & 0.73 \\ 
& LLM Expand & 0.66 & 0.678 & 0.663 & 0.681 & 0.67 \\ 
& Back-Translate & 0.57 & 0.571 & 0.57 & 0.570 & 0.57 \\ 
& Paraphrase & 0.55 & 0.588 & 0.55 & 0.568 & 0.55 \\ 
& Text Generate & 0.52 & 0.63 & 0.52 & 0.569 & 0.52 \\

\hline
\multirow{3}{4em}{Sindhi} & Baseline & 0.93 & 0.931& 0.93 & 0.929 & 0.93 \\ 
& Synonym Replacement & 0.88 & 0.885 & 0.879& 0.882 & 0.88 \\ 
& Random Insertion & 0.79 & 0.827 & 0.79 & 0.808 & 0.79 \\ 
& Random Swap & 0.96 & 0.962 & 0.96 & 0.959 & 0.96 \\ 
& Random Delete & 0.99 & 0.990 & 0.99 & \textbf{0.989} & 0.99 \\ 
& LLM Expand & 0.98 & 0.985 & 0.98 & 0.985 & 0.98 \\ 
& Back-Translate & 0.96 & 0.960 & 0.96 & 0.96 & 0.96 \\ 
& Paraphrase & 0.94 & 0.94 & 0.94 & 0.94 & 0.94 \\ 
& Text Generate & 0.95 & 0.951 & 0.95 & 0.950 & 0.95 \\

\hline
\end{tabular}
\caption{\label{citation-guide}
Results for the task of Binary Classification
}
\end{center}
\end{table*}

\begin{table*}
\centering
\begin{center}
\begin{tabular}{ lllllll } 
\hline
\textbf{Language} & \textbf{Methodology} & \textbf{Accuracy} & \textbf{Precision}	& \textbf{Recall} & \textbf{F1 Macro} & \textbf{F1 Mirco} \\

\hline
\multirow{3}{4em}{Sanskrit} & Baseline & 0.71 & 0.749 & 0.68 & 0.616 & 0.71 \\ 
& Synonym Replacement & 0.79 & 0.8072 & 0.808 & \textbf{0.807 } & 0.79 \\ 
& Random Insertion & 0.78 & 0.787 & 0.777 & 0.782 & 0.78 \\ 
& Random Swap & 0.7 & 0.523 & 0.666 & 0.575 & 0.7 \\ 
& Random Delete & 0.77 & 0.775 & 0.772 & 0.773 & 0.77 \\ 
& LLM Expand & 0.77 & 0.778 & 0.780 & 0.778 & 0.77 \\ 
& Back-Translate & 0.71 & 0.712 & 0.705 & 0.708 & 0.71 \\ 
& Paraphrase & 0.74 & 0.748 & 0.75 & 0.749 & 0.74 \\ 
& Text Generate & 0.75 & 0.759 & 0.741 & 0.750 & 0.75 \\

\hline
\multirow{3}{4em}{Marathi} & Baseline & 0.298 & 0.247 & 0.298 & 0.261 & 0.298 \\ 
& Synonym Replacement & 0.432 & 0.422 & 0.432 & \textbf{0.427} & 0.432 \\ 
& Random Insertion & 0.394 & 0.396 & 0.394 & 0.395 & 0.394 \\ 
& Random Swap & 0.278 & 0.187 & 0.278 & 0.191 & 0.278 \\ 
& Random Delete & 0.365 & 0.376 & 0.365 & 0.360 & 0.365 \\ 
& LLM Expand & 0.361 & 0.366 & 0.360 & 0.367 & 0.361 \\ 
& Back-Translate & 0.480 & 0.373 & 0.480 & 0.420 & 0.480 \\ 
& Paraphrase & 0.326 & 0.376 & 0.326 & 0.349 & 0.326 \\ 
& Text Generate & 0.326 & 0.319 & 0.326 & 0.323 & 0.326 \\

\hline
\multirow{3}{4em}{Hindi} & Baseline & 0.53 & 0.407 & 0.53 & 0.455 & 0.53 \\ 
& Synonym Replacement & 0.55 & 0.516 & 0.55 & 0.532 & 0.55 \\ 
& Random Insertion & 0.58 & 0.618 & 0.58 & \textbf{0.598} & 0.58 \\ 
& Random Swap & 0.52 & 0.464 & 0.52 & 0.484 & 0.52 \\ 
& Random Delete & 0.58 & 0.568 & 0.58 & 0.549 & 0.58 \\ 
& LLM Expand & 0.59 & 0.592 & 0.59 & 0.581 & 0.59 \\ 
& Back-Translate & 0.56 & 0.537 & 0.56 & 0.548 & 0.56 \\ 
& Paraphrase & 0.53 & 0.5132 & 0.51 & 0.521 & 0.53 \\ 
& Text Generate & 0.51 & 0.5362 & 0.51 & 0.522 & 0.51 \\

\hline
\multirow{3}{4em}{Gujarati} & Baseline & 0.47 & 0.480 & 0.469 & 0.475 & 0.47 \\ 
& Synonym Replacement & 0.51 & 0.382 & 0.506 & 0.435 & 0.51 \\ 
& Random Insertion & 0.68 & 0.684 & 0.679 & 0.681 & 0.68 \\ 
& Random Swap & 0.68 & 0.713 & 0.679 & 0.666 & 0.68 \\ 
& Random Delete & 0.78 & 0.786 & 0.780 & \textbf{0.779} & 0.78 \\ 
& LLM Expand & 0.57 & 0.573 & 0.58 & 0.572 & 0.57 \\ 
& Back-Translate & 0.58 & 0.585 & 0.581 & 0.583 & 0.58 \\ 
& Paraphrase & 0.54 & 0.544 & 0.54 & 0.54 & 0.54 \\ 
& Text Generate & 0.62 & 0.646 & 0.62 & 0.633 & 0.59 \\

\hline
\multirow{3}{4em}{Telugu} & Baseline & 0.45 & 0.476 & 0.473 & 0.488 & 0.46 \\ 
& Synonym Replacement & 0.72 & 0.754 & 0.719 & 0.736 & 0.72 \\ 
& Random Insertion & 0.72 & 0.718 & 0.718 & 0.718 & 0.72 \\ 
& Random Swap & 0.67 & 0.728 & 0.674 & 0.663 & 0.68 \\ 
& Random Delete & 0.78 & 0.792 & 0.792 & \textbf{0.782} & 0.78 \\ 
& LLM Expand & 0.57 & 0.578 & 0.58 & 0.581 & 0.57 \\ 
& Back-Translate & 0.878 & 0.885 & 0.877 & 0.881 & 0.878 \\ 
& Paraphrase & 0.646 & 0.699 & 0.647 & 0.672 & 0.646 \\ 
& Text Generate & 0.747 & 0.773 & 0.747 & 0.760 & 0.747 \\

\hline
\multirow{3}{4em}{Sindhi} & Baseline & 0.49 & 0.492 & 0.489 & 0.489 & 0.49 \\ 
& Synonym Replacement & 0.48 & 0.671 & 0.475 & 0.556 & 0.48 \\ 
& Random Insertion & 0.9 & 0.907 & 0.9 & \textbf{0.904} & 0.9 \\ 
& Random Swap & 0.57 & 0.574 & 0.568 & 0.558 & 0.57 \\ 
& Random Delete & 0.89 & 0.896 & 0.8903 & 0.890 & 0.89 \\ 
& LLM Expand & 0.76 & 0.783 & 0.7624 & 0.758 & 0.787 \\ 
& Back-Translate & 0.72 & 0.755 & 0.722 & 0.738 & 0.72 \\ 
& Paraphrase & 0.9 & 0.906 & 0.906 & 0.903 & 0.9 \\ 
& Text Generate & 0.63 & 0.63 & 0.626 & 0.516 & 0.63 \\

\hline
\end{tabular}
\caption{\label{citation-guide2}
Results for the task of Multiclass Classification
}
\end{center}
\end{table*}

\section{Results}
For Binary text classification, As seen in Table \ref{citation-guide}, all the data augmentation tasks for Sanskrit perform better than the baseline model, with Synonym replacement and random deletion being the best performers. Similarly, for Marathi, all the models exceed the performance of the baseline model, with Translate and Generate modules being the best performer. Similar trends of the augmentations models outperforming the baseline models are observed in every language except Hindi, where the baseline model itself generates a very high score for the very little data it was fine-tuned with. This increase in performance can be attributed to the availability of a large amount of Hindi data on the internet. The abundance of Hindi text data allows the model to learn from a vast range of unannotated text, which helps build strong language representations and capture relevant linguistic patterns. Apart from very few exceptions in Sindhi and Telugu text classification tasks, the augmented models perform better than the baseline model. 

Similarly, for the multiclass text classification task, the augmentation methods outperform the baseline methods for every language with very few exceptions. Many augmentation techniques perform better than the baseline model and the binary classification task. Detailed results are given in Table  \ref{citation-guide2}.

We can see from both the result tables that basic augmentation using Easy Data Augmentation outperforms advanced augmentation techniques like augmentation, Paraphrasing, Summarization, and generation.

\section{Error analysis}

We did not really encounter any failure cases in our analysis. The datasets we used were highly cited/well-curated ones, and fine-tuning HuggingFace's multi-lingual BERT was error-free as well.

The random Insertion technique for EDA works on the phenomenon that we randomly insert a synonym of a word(the selected word should not be a stop word). For some languages like Gujarati, Sindhi, and Sanskrit there is no available list of stop words that we can exclude from the sentence. Hence, at times it is possible that a stopword might be replaced with a synonym which in turn can reduce the accuracy. This can be seen in binary classification in the Sindhi language. Here the F1 score drops from 0.929 to 0.008.

Due to the generation limit of GPT, augmentation techniques which used Large Language Models like GPT,  were incomplete when big sentences were passed as input. This is especially seen in the performance of the Text Generation module for the Telugu Binary Text Classification Task where no substantial model scores are obtained. The average length of the augmented examples was found to be lesser than the average length of the original dataset. For example, for binary Telugu Text Classification, the average length of augmented samples was 46, whereas the average length of the original dataset was 72. This is one of the reasons for the lack of improvement.

\section{Conclusion}
From the results, we can see that the proposed eight methodologies for data augmentation consistently outperform the baseline on both binary and multiclass classification tasks across all the languages. Amongst these methods, EDA is a clear winner showing consistently great performance across all languages. For the task of binary classification, random delete, LLM Expand, and text generate methodologies show good performance overall as compared to other methods. A similar trend can also be observed in the case of multiclass classification, however, even Back-Translate demonstrates good performance here. We were surprised to see the almost consistent good performance demonstrated by the random delete methodology because at the face of it, random delete seems to be eliminating information from each sentence instead of supplementing it.

\section*{Limitations}
We have limited the scope of our research work to a specific set of languages due to the unavailability of word embeddings for most of the other Indian languages. We plan to explore more augmentation techniques in future work. Due to the limitation of computational resources we could not use better Large Language Models.

\section*{Ethics Statement}

The dataset from this work may contain some harmful words due to the fact that we are working on Hate Detection Task. We do not support the use of such words in day-to-day life. We have abided by the ACL Ethics Policy while researching and writing the paper. This work is novel and not plagiarised from another source. 



\bibliography{main}
\bibliographystyle{acl_natbib}

\end{document}